\pdfoutput=1

\documentclass[10pt,twocolumn,letterpaper]{article}
\usepackage[pagenumbers]{cvpr} 

\usepackage{amsfonts}
\usepackage{adjustbox}
\usepackage{placeins}

\definecolor{cvprblue}{rgb}{0.21,0.49,0.74}
\usepackage[pagebackref,breaklinks,colorlinks,allcolors=cvprblue]{hyperref}

\def\confName{CVPR}
\def\confYear{2027}

\title{StyleID++: Position-Decoupled and Structure-Coherent Style Retrieval across Diffusion Architectures}

\author{Chun Xie\\
University of Tsukuba\\
{\tt\small xie.chun@image.iit.tsukuba.ac.jp}
\and
Itaru Kitahara\\
University of Tsukuba\\
{\tt\small kitahara.itaru@image.iit.tsukuba.ac.jp}
}

\begin{document}
\maketitle
\begin{abstract}
Attention injection offers a remarkably simple route to training-free style transfer: content queries retrieve style keys and values inside a pretrained diffusion model.
We show that this operation also creates an overlooked failure mode.
Position signals that are useful within one image become a spurious shared coordinate system across two spatially unrelated images, causing content tokens to favor style tokens at similar coordinates.
This \emph{cross-image positional leakage} appears in both convolutional U-Nets and diffusion transformers (DiTs), despite their different position mechanisms.
Removing the positional shortcut exposes a second problem: unconstrained global retrieval lets related content tokens select incompatible style regions, producing fragmented appearance.
We present StyleID++, a training-free framework that addresses the two problems jointly.
Position-Decoupled Retrieval reduces coordinate-dependent matching through circular padding in U-Nets and injection-only RoPE removal in DiTs, while preserving native position modeling elsewhere.
Affinity-Coherent Refinement then regularizes style assignments with the content self-attention graph, encouraging structurally related tokens to retrieve compatible style evidence.
We quantify the leakage with a coordinate--centroid correlation and a
permutation test on the style reference.
The permutation error can be driven to zero in both backbones, which
identifies the position mechanism as the carrier of the leakage, and
StyleID++ obtains the best ArtFID of all methods we compare.
\end{abstract}

\section{Introduction}
\label{sec:introduction}

\begin{figure*}[tb]
    \centering
    \includegraphics[width=\textwidth]{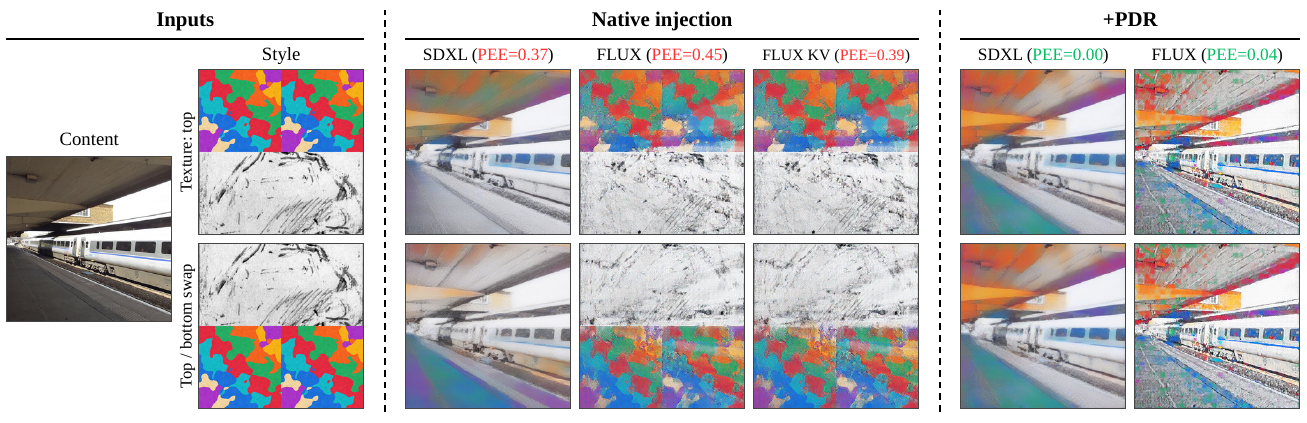}
    \caption{\textbf{Cross-image positional leakage under a top/bottom swap.}
    For a fixed content image, exchanging the two halves of the style reference
    moves the colored texture from top to bottom. FLUX KV applies the swap
    to feature token positions from the original style image instead of pixel
    regions. Native SDXL injection
    relocates the color response, while FLUX and FLUX KV strongly follow the
    reference layout. Position-Decoupled Retrieval (PDR) reduces this
    sensitivity while retaining the content structure. PEE values describe
    this example, whereas \cref{tab:leakage} reports dataset averages.
    The supplement includes all three swap types.}
    \label{fig:teaser}
\end{figure*}

Training-free diffusion stylization methods such as StyleID~\cite{styleid}
inject style by retaining content queries while replacing self-attention keys
and values with those recorded from a style image.
The resulting all-to-all attention appears to let each content token retrieve
style evidence from anywhere in the reference.
What happens, then, if the upper and lower halves of the style image are
exchanged before injection?
Ideally, the retrieved evidence should move with the style patterns.
Instead, the colored regions in the outputs shift with the relocated texture,
despite the unchanged content image and local style patterns
(\cref{fig:teaser}).
This indicates that retrieval depends on where style evidence is placed,
rather than solely on its appearance.
The same behavior occurs in a convolutional Stable Diffusion U-Net and a
transformer-based FLUX model.
This simple intervention reveals that attention-based style transfer is not
performing the unrestricted content--style matching suggested by its
all-to-all formulation.

The failure becomes clear once feature injection is viewed as
\emph{cross-image retrieval}: every content token uses attention to select
style evidence from a second image.
This view exposes a mismatch between pretraining and inference.
Position signals are beneficial when queries and keys describe one image, but
content and style images generally do not share a coordinate system.
When the same position mechanisms are retained after cross-image injection,
they create a shortcut that favors matching coordinates regardless of where
the relevant appearance lies.
We call this failure \emph{cross-image positional leakage}.

Importantly, the leakage is not tied to one architectural primitive.
In a convolutional U-Net, repeated zero padding supplies boundary-dependent
position cues~\cite{paddingbias}.
In a DiT, Rotary Position Embeddings (RoPE) explicitly modulate query--key
similarity by relative coordinates.
Both are valid spatial priors for within-image generation, yet both become
misplaced when independently arranged images are coupled by attention.
Thus, the shared failure lies not in a particular position operator, but in
carrying within-image coordinate priors into cross-image retrieval, where the
two images have unrelated layouts.

Suppressing this shortcut is necessary but insufficient.
Once a query can retrieve from the entire style image without coordinate
preference, each token makes an almost independent choice.
Related content tokens may then select disconnected or incompatible style
regions, yielding mottled colors, broken strokes, and unstable textures.
The problem therefore changes from \emph{where a token is allowed to look} to
\emph{how retrieval decisions should agree}.
This is not conventional semantic correspondence, which assumes aligned
object categories in the content and style images.
Rather, arbitrary style transfer requires an annotation-free structural prior:
tokens that strongly support one another in the content representation should
draw compatible evidence from the style image, even when the two images depict
different objects.

We address this coupled problem with StyleID++.
Our first component, \emph{Position-Decoupled Retrieval} (PDR), targets the
backbone-specific position source that contaminates cross-image matching.
For U-Nets, we replace zero padding with circular padding during inversion and
sampling to weaken boundary-anchored coordinates.
For DiTs, we compute the injected content-query/style-key logits without RoPE
while retaining RoPE for the model's native within-image operations.
Our second component, \emph{Affinity-Coherent Refinement} (ACR), constructs an
affinity graph from content self-attention and uses it to coordinate the
candidate style keys of related queries.
Both components are training-free and require neither labels nor an auxiliary
model.

Our contributions are threefold:
\begin{itemize}
    \item We identify and measure cross-image positional leakage, a shared
    failure of StyleID-like injection across U-Net and DiT backbones, and
    formulate a permutation test that distinguishes appearance retrieval from
    coordinate following.
    \item We introduce a backbone-aware position-decoupling principle:
    neutralize implicit boundary cues in U-Nets with circular padding, and remove RoPE only from the
    injected matching path in DiTs, without discarding the position modeling
    needed by the native generator.
    \item We expose the disorder created by unconstrained global retrieval and
    propose content-affinity refinement to make style assignments coherent.
\end{itemize}

\section{Related Work}

\paragraph{Training-free diffusion style transfer.}
Diffusion models~\cite{ho_ddpm,rombach_ldm} provide strong generative priors
for transferring appearance from reference images.
One family encodes style as an auxiliary condition or manipulates the
initialization and sampling trajectory
~\cite{instantstyle,instantstyle_plus,styleshot,stylestudio,stylessp}.
A complementary family directly edits internal attention.
StyleID~\cite{styleid} replaces content keys and values with style features.
StyleAligned~\cite{StyleAlign} shares attention statistics across generation
branches. DiffStyle~\cite{Diffstyle} manipulates diffusion features without
per-image training.
These methods demonstrate the effectiveness of inference-time injection, but
mainly study how much and when to inject.
We instead examine the correspondence rule induced when within-image
self-attention is repurposed across independently arranged images.

\paragraph{Regional coherence in attention-based stylization.}
Patch-wise attention has long been used to match content and style
features~\cite{park_sanet,liu_adaattn,deng_stytr}.
Recent methods improve consistency using key-region selection, global style
representations, learned semantic regions, or auxiliary constraints
~\cite{a2k,scsa,hsi,stylegallery}.
SCSA~\cite{scsa}, for example, restricts queries to continuous and sparse keys
within corresponding semantic regions, while StyleGallery~\cite{stylegallery}
clusters and matches regions before optimization.
In contrast, we consider arbitrary content--style pairs without assuming
category correspondence, masks, or a learned region matcher.
Our method uses only the content model's own affinity graph to regularize
otherwise independent cross-image retrieval decisions.

\paragraph{Position signals in visual generation and stylization.}
Position encodings are essential spatial inductive biases, but their form can
affect translation behavior and generalization.
Zero padding can act as an implicit coordinate system in convolutional
generators~\cite{paddingbias}, while transformer style transfer has motivated
content-aware alternatives to standard position encoding~\cite{deng_stytr}.
DiTs introduce explicit position mechanisms, including RoPE, into generative
attention~\cite{sd3,flux}.
Prior work studies how these signals represent layout within an image.
We study a different boundary condition: a position prior learned for
within-image attention is carried into attention between two images whose
coordinates have no prescribed correspondence.

\section{Method}
\label{sec:method}

\subsection{Preliminaries}
\label{sec:retrieval}

\paragraph{Style injection as cross-image retrieval.}

At a selected block and denoising step, let
$Q_c\in\mathbb{R}^{N_c\times d}$ be the content queries and
$K_s,V_s\in\mathbb{R}^{N_s\times d}$ be the recorded style keys and values.
StyleID-like injection retains the content queries and substitutes the style
keys and values:
\begin{equation}
\begin{aligned}
  L_{cs}&=\frac{Q_cK_s^\top}{\sqrt d},&
  A_{cs}&=\operatorname{softmax}(L_{cs}),\\
  Y_c&=A_{cs}V_s.&&
\end{aligned}
  \label{eq:injection}
\end{equation}
Each row of $A_{cs}$ is therefore a retrieval distribution over style tokens.

\subsection{Diagnosing Cross-Image Positional Leakage}
\label{sec:leakage}

Although every style key is reachable in \cref{eq:injection}, the match is not
purely appearance-based: both $Q_c$ and $K_s$ are produced by the pretrained
backbone and inherit its position cues.
These cues are useful for modeling geometry within one image.
Across independently arranged content and style images, however, a preference
for similar coordinates becomes an irrelevant retrieval shortcut.

We test this shortcut with a known grid-aligned rearrangement $T$, such as a half- or quadrant-swap.
Let $\pi_T(j)$ denote the new location of original style token $j$ after
applying $T$.
We transform the \emph{input style image} and rerun the same style branch,
rather than permuting cached keys:
\[
\begin{aligned}
 K_s^{(T)}&=\Phi_K(T(x_s)),\\
 A_{cs}^{(T)}&=\operatorname{softmax}
 \left(Q_c(K_s^{(T)})^\top/\sqrt d\right).
\end{aligned}
\]
where $\Phi_K$ includes style inversion and key recording up to the evaluated
block and step, with $K_s=\Phi_K(x_s)$ for the original reference.
We then align the transformed attention back to the original style-token
order by simple column indexing,
$[\mathcal U_T(A_{cs}^{(T)})]_{i,j}
=[A_{cs}^{(T)}]_{i,\pi_T(j)}$.
If retrieval follows the moved style evidence, the aligned distribution
$\mathcal U_T(A_{cs}^{(T)})$ should match $A_{cs}$.
We measure their discrepancy with the \emph{Permutation Equivariance Error}
(PEE):
\begin{equation}
  \operatorname{PEE}(T)=\frac{1}{N_c}
  \sum_{i=1}^{N_c}\operatorname{JS}
  \left(A_{cs}^{i,:},
  \mathcal U_T(A_{cs}^{(T)})^{i,:}\right),
  \label{eq:pee}
\end{equation}
where $\operatorname{JS}$ is the Jensen--Shannon divergence between two
retrieval distributions.
PEE is zero under exact equivariance and increases when attention remains
attached to the old coordinates.
The main test uses quadrant swaps aligned with the token grid, while the
figures use the simpler upper/lower half swap for visual clarity.
As a complementary diagnostic on an unmodified style reference, let
$\mathbf p_i^c=(x_i^c,y_i^c)$ and
$\mathbf p_j^s=(x_j^s,y_j^s)\in[0,1]^2$ be the normalized coordinates of
content query $i$ and style key $j$, respectively.
For every content query, we first compute the attention-weighted centroid of
the retrieved style locations,
\begin{equation}
\bar{\mathbf p}_i^s=(\bar x_i^s,\bar y_i^s)
=\sum_{j=1}^{N_s}A_{cs}(i,j)\mathbf p_j^s.
\label{eq:centroid}
\end{equation}
We then measure whether queries at larger horizontal or vertical coordinates
also retrieve style evidence at larger corresponding coordinates:
\begin{equation}
\operatorname{Corr}(A_{cs})
=\frac{
\left|\rho_{\mathrm S}(\{x_i^c\},\{\bar x_i^s\})\right|
+
\left|\rho_{\mathrm S}(\{y_i^c\},\{\bar y_i^s\})\right|
}{2},
\label{eq:corr}
\end{equation}
where $\rho_{\mathrm S}$ denotes Spearman's rank correlation and each set
runs over the $N_c$ content queries.
We report the mean Corr over the evaluated attention heads, blocks, denoising
steps, and image pairs.
Thus, Corr measures the coordinate-dependent preference on ordinary
inputs, while PEE measures the failure of attention to follow deliberately
moved style evidence.

\begin{figure*}[tb]
    \centering
    \includegraphics[width=0.65\linewidth]{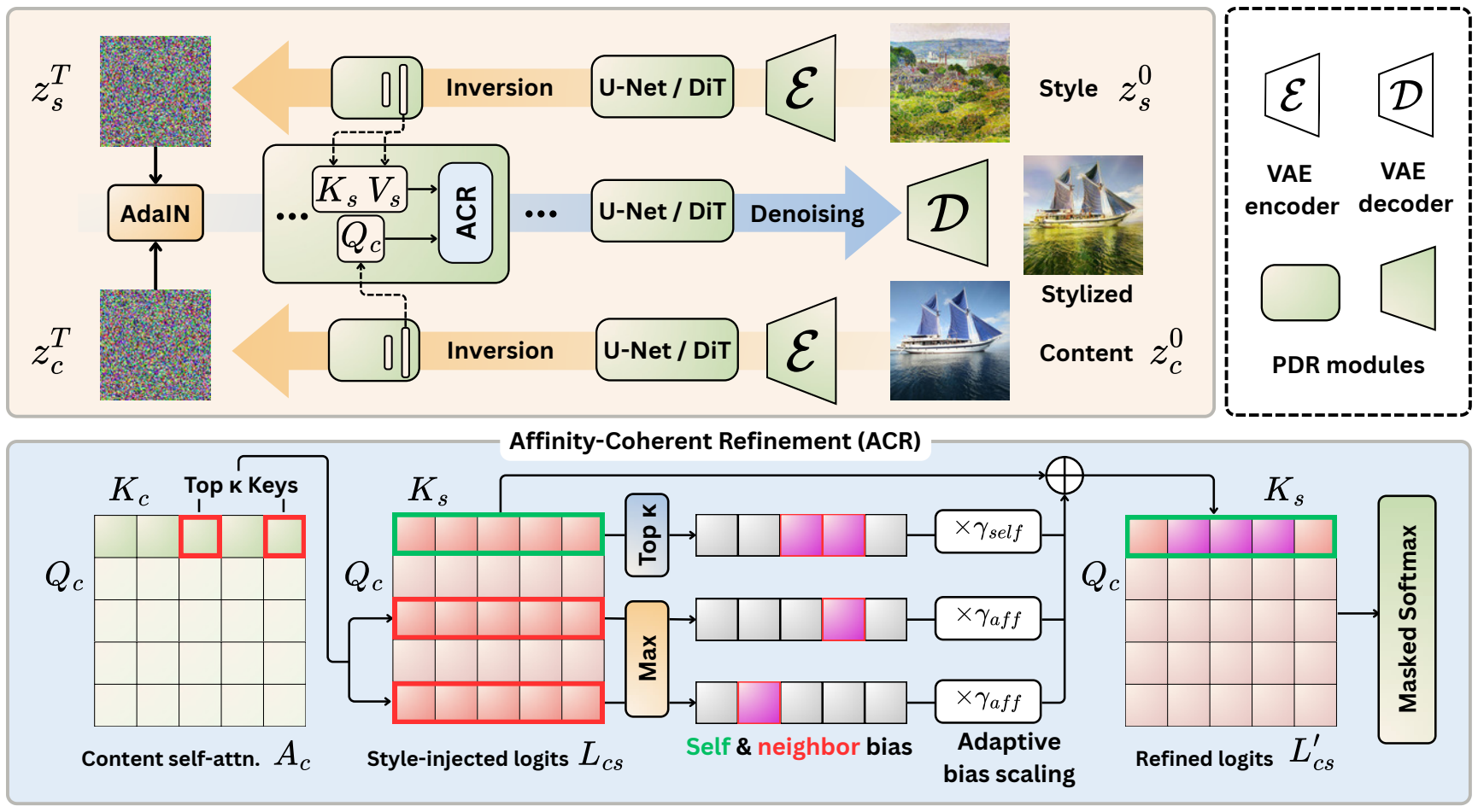}
    \caption{
    Overview of our StyleID++ framework. \textbf{Top:} The standard Style Injection via Attention pipeline used in StyleID, where latent features from inversion of the content and style images are injected into the attention layers during denoising. We apply the Position-Decoupled Retrieval (PDR) to all modules to remove positional bias. \textbf{Bottom:} ACR. The content self-attention map supplies the strongest
    neighbors of each query. The query's own top-scoring style keys and the
    dominant key of each neighbor receive separate biases, which are scaled by
    the spread of the row's logits.}
    \label{fig:overview}
\end{figure*}

\subsection{StyleID++ Overview}
StyleID++ separates two roles that are conflated by direct injection:
position modeling within each image and correspondence between images.
PDR reduces the position dependence inherited by cross-image retrieval, while
ACR constrains the larger retrieval space exposed by PDR.

\subsection{Position-Decoupled Retrieval}
\label{sec:pdr}

\paragraph{U-Net: neutralizing implicit boundary anchors.}
Repeated zero padding makes distances to image boundaries recoverable from
convolutional features~\cite{paddingbias}.
We replace zero padding in the VAE and denoising U-Net with circular padding
during both inversion and sampling.
For $F\in\mathbb{R}^{H\times W\times C}$ and padding width $r$, the padded
feature is defined on the extended index range
$-r\le i<H+r$ and $-r\le j<W+r$ by
\begin{equation}
 F_{\mathrm{circ}}[i,j]=F[i\bmod H,\;j\bmod W],
\end{equation}
so that positions outside the feature map wrap around.

This operation does not make the representation position-free, nor does it
change the mathematical receptive field of attention.
It specifically removes the constant boundary values that provide a strong
absolute-coordinate anchor, thereby reducing their influence on
cross-image logits.
Reflection and replication padding are evaluated as causal controls.

\paragraph{DiT: decoupling RoPE only in the injected path.}
Let $R(p)$ be the RoPE rotation at coordinate $p$.
Native attention uses
\begin{equation}
  L_{\mathrm{native}}(i,j)=
  \frac{(R(p_i)q_i)^\top(R(p_j)k_j)}{\sqrt d},
  \label{eq:rope}
\end{equation}
whose relative-position term is appropriate when $i$ and $j$ belong to the
same image.
Applying it unchanged to content queries and style keys imposes a relation
between coordinates from different images.
PDR instead computes the injected logits as
\begin{equation}
  L_{\mathrm{inj}}^{\mathrm{PDR}}(i,j)
  =\frac{q_c^i{}^\top k_s^j}{\sqrt d}.
  \label{eq:ditpdr}
\end{equation}
All native content, style, and generation paths retain RoPE.
Thus PDR removes the cross-image coordinate coupling without depriving the
DiT of the geometry needed to synthesize coherent images.

\subsection{Affinity-Coherent Refinement}
\label{sec:acr}

\paragraph{From independent matching to retrieval consensus.}
PDR removes the coordinate shortcut, but \cref{eq:injection} still treats the
style assignment of every content query as an independent decision.
The enlarged search space often contains several near-tied style modes, so
small logit differences can make strongly related content tokens retrieve
incompatible patterns.
Our guiding principle is therefore not that related tokens must select an
identical style key, but that their retrieval decisions should share
compatible evidence.

We use $L_{cs}$ to denote the cross-image logits after PDR.
We obtain an annotation-free relation graph from native content self-attention
recorded at the same block and timestep \emph{before} style injection:
\begin{equation}
  A_c=\operatorname{softmax}
  \left(\frac{Q_cK_c^\top}{\sqrt d}\right).
  \label{eq:contentaffinity}
\end{equation}
We call these relations \emph{affinities} rather than semantic labels:
self-attention can connect object parts, repeated structures, or context.
For query $i$, let
$\mathcal{N}_i=\operatorname{TopK}(A_c^{i,:},\kappa_q)$ be its most strongly
related content tokens.
Let $\mathcal{T}_i=\operatorname{TopK}(L_{cs}^{i,:},\kappa_s)$ denote its own
high-scoring style candidates, and let
$m_j=\arg\max_k L_{cs}(j,k)$ be the dominant style key of neighbor $j$.
ACR augments the query-supported candidates with these one-hop neighbor modes:
\begin{equation}
 \mathcal{C}_i=\mathcal{T}_i\cup\{m_j\mid j\in\mathcal{N}_i\}.
 \label{eq:candidates}
\end{equation}
Define the unary support
$u_i(k)=\mathbb{I}[k\in\mathcal{T}_i]$ and the affinity support
$g_i(k)=\mathbb{I}[\exists j\in\mathcal{N}_i:m_j=k]$.
The binary $g_i$ is a clipped graph message: multiple neighbors proposing the
same key do not increase its score without bound.
This makes ACR a sparse one-step message-passing rule in \emph{style-retrieval
space}, rather than spatial smoothing in the content image.

\paragraph{Calibrated unary and relational evidence.}
We refine the retained logits as
\begin{equation}
 L'_{cs}(i,k)=L_{cs}(i,k)
 +\gamma_{\mathrm{self}}^i u_i(k)
 +\gamma_{\mathrm{aff}}^i g_i(k),
\label{eq:refinelogits}
\end{equation}
where the two terms play different roles.
$\gamma_{\mathrm{self}}^i$ preserves evidence selected by query $i$ itself,
preventing neighbor messages from overwriting a confident individual match.
$\gamma_{\mathrm{aff}}^i$ promotes modes supported by related content tokens,
providing the coherence pressure missing from independent retrieval.
A key supported by both sources receives both bonuses, so agreement between
local evidence and the affinity graph is rewarded most strongly.

Raw logit dispersion varies substantially across blocks, timesteps, and
backbones.
A fixed additive bias would therefore be negligible for a sharp row but
dominant for a flat one.
We instead measure both biases in units of the row's own logit standard
deviation:
\begin{equation}
\begin{aligned}
 \sigma_i&=\operatorname{std}(L_{cs}^{i,:}),\\
 \gamma_{\mathrm{self}}^i
 &=\lambda_{\mathrm{self}}\sigma_i,\qquad
 \gamma_{\mathrm{aff}}^i
 =\lambda_{\mathrm{aff}}\sigma_i.
\end{aligned}
\label{eq:scaling}
\end{equation}
Equivalently,
$(L'_{cs}(i,k)-L_{cs}(i,k))/\sigma_i
=\lambda_{\mathrm{self}}u_i(k)+\lambda_{\mathrm{aff}}g_i(k)$.
Thus, $\lambda_{\mathrm{self}}$ and $\lambda_{\mathrm{aff}}$ are
dimensionless decision margins: the former controls fidelity to a query's own
ranking, while the latter controls how much affinity evidence is needed to
resolve a near tie.

Finally,
\begin{equation}
 A'_{cs}(i,k)=
 \begin{cases}
 \displaystyle
 \frac{\exp L'_{cs}(i,k)}
 {\sum_{\ell\in\mathcal{C}_i}\exp L'_{cs}(i,\ell)},
 & k\in\mathcal{C}_i,\\[6pt]
 0, & \text{otherwise}.
 \end{cases}
\label{eq:maskedsoftmax}
\end{equation}
For example, if content tokens $q_1$ and $q_2$ have strong affinity and
$q_1$ retrieves style mode $s_1$, then $s_1$ enters $\mathcal{C}_2$ through
$g_2$.
If $s_1$ is also plausible under $q_2$'s own logits, it receives the
self--affinity consensus bonus. If $q_2$ has stronger incompatible evidence,
its own top candidates remain available.
ACR therefore encourages coherent retrieval without forcing all related
queries to copy one assignment.

\section{Experiments}
\label{sec:experiments}

\subsection{Experimental Protocol}

\paragraph{Backbones and data.}
We evaluate a convolutional SDXL backbone~\cite{podell2024sdxl} and a
transformer-based FLUX backbone~\cite{flux}.
For SDXL, we use DDIM inversion and sampling~\cite{song_ddim} with 20 steps.
For FLUX, we use Euler inversion and sampling with 14 steps.
Following the evaluation protocol of StyleSSP~\cite{stylessp}, we use the
public test set released by InstantStyle-Plus~\cite{instantstyle_plus}, which
contains 20 MS-COCO content images~\cite{lin2014coco} and 103 WikiArt style
images~\cite{wikiart}.
We evaluate all possible content--style combinations, producing 2,060
stylized outputs in total.

\paragraph{Implementation details.}
Within each experiment, generation settings are matched across the compared conditions for each backbone. The complete configuration, including injection blocks, denoising intervals, $\kappa_q$, $\kappa_s$, $\lambda_{\rm self}$, and
$\lambda_{\rm aff}$, is provided in the supplementary material.
All experiments are run on a single NVIDIA H100 GPU.

\paragraph{Metrics.}
We report ArtFID~\cite{artfid2023}, FID~\cite{heusel2017fid}, and
LPIPS~\cite{zhang2018lpips} for comparison with prior work.
LPIPS measures the distance between a stylized result and its content image,
FID measures the distributional alignment between the stylized outputs and the
style images, and ArtFID combines the two as
$\mathrm{ArtFID}=(1+\mathrm{LPIPS})(1+\mathrm{FID})$.
Mechanism-specific evaluation uses the coordinate--centroid correlation
(Corr, \cref{eq:corr}), the Permutation Equivariance Error
(PEE, \cref{eq:pee}), and LPIPS$_{\mathrm{swap}}$, the LPIPS between the
outputs obtained from the original and the rearranged style reference.
We additionally introduce Contrastive Style Evidence Consistency (CSEC@8, \cref{sec:acr_effect}) to measure whether agreement in retrieved style evidence reflects content similarity.

\subsection{Positional Leakage across Backbones}

We instantiate the Corr and PEE diagnostics with grid-aligned quadrant swaps
of the style reference.
Both backbones show a strong coordinate dependence under the injection
baseline, and PDR substantially reduces it on each
(\Cref{tab:leakage}).
That two backbones with unrelated position mechanisms behave alike supports
our account: the leakage follows from carrying within-image position priors
into cross-image retrieval, not from any particular position operator.

\begin{table}[tb]
\centering
\caption{\textbf{Cross-image positional leakage.}
LPIPS$_{\mathrm{swap}}$ is measured between the outputs from the original
and the quadrant-swapped style reference, not against the content image.}
\label{tab:leakage}
\begin{adjustbox}{width=\columnwidth}
\begin{tabular}{llccc}
\toprule
Backbone & Retrieval & Corr.$\downarrow$ & PEE$\downarrow$ & LPIPS$_{\mathrm{swap}}\downarrow$ \\
\midrule
SDXL & Baseline & 0.7987            & 0.5059            & 0.1227\\
SDXL & PDR      & \textbf{0.3796}   & \textbf{0.0000}   & \textbf{0.0001}\\
\midrule
FLUX & Baseline & 0.7727             & 0.5665            & 0.6346\\
FLUX & PDR      & \textbf{0.1754}    & \textbf{0.0748}   & \textbf{0.0018}\\
\bottomrule
\end{tabular}
\end{adjustbox}
\end{table}

\subsection{Isolating the Position Mechanism}

\begin{figure}[tb]
    \centering
    \includegraphics[width=\linewidth]{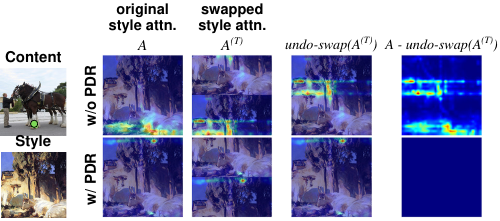}
    \caption{\textbf{Content→style attention under a top/bottom swap of the style image.}
    Using the same content query (green dot), attention over style keys is recomputed after swapping the two halves of the style image, then aligned back by undoing the known permutation. Without PDR (top), attention stays anchored to absolute coordinates and undoing the swap fails to recover the original map. With PDR (bottom), the attention translates exactly with the style content and the aligned map matches the original, leaving a flat residual (PEE=0). \textit{Further visualizations are provided in the supplementary material.}}
    \label{fig:swap}
\end{figure}

Reducing Corr and PEE alone does not establish position decoupling as the
cause, since padding and RoPE changes also affect other backbone behavior. We therefore vary one position
mechanism at a time in \cref{tab:position_ablation} and inspect individual
queries under the style swap in \cref{fig:swap}. Reflection and replication
padding remove the boundary constant and cut PEE roughly fivefold,
indicating that this constant, rather than padding itself, carries implicit
coordinates. Circular padding additionally makes the style branch exactly
shift-equivariant. Removing RoPE throughout FLUX eliminates coordinate
preference, confirming its role, but serves as a causal check rather than
the proposed configuration. Native RoPE achieves the lowest FID only at the cost
of poor content preservation (LPIPS $0.7705$). Injection-only removal gives
a better trade-off: global removal further preserves content (lower LPIPS)
but matches the style distribution less well (higher FID). Together, these
results support separating within-image position modeling from cross-image
correspondence.

\begin{table}[tb]
\centering
\caption{\textbf{Position mechanism ablation.}
SDXL rows vary only the padding mode, FLUX rows only the place where RoPE is
applied. Circular padding attains zero PEE and the best score on all three
quality metrics. For FLUX, injection-only RoPE removal gives the best ArtFID,
while global removal achieves near-zero PEE but higher FID than injection-only removal.}
\label{tab:position_ablation}
\begin{adjustbox}{width=\columnwidth}
\begin{tabular}{llcccc}
\toprule
Backbone & Position treatment & PEE$\downarrow$ & ArtFID$\downarrow$ &FID$\downarrow$ & LPIPS$\downarrow$\\
\midrule
SDXL & Zero padding     & 0.5059    & 22.489    & 14.190    & 0.4805\\
SDXL & Reflection       & 0.1284    & 21.382    & 13.469    & 0.4778\\
SDXL & Replication      & 0.1327    & 21.337    & 13.433    & 0.4783\\
SDXL & Circular         & \textbf{0.0000} & \textbf{21.198}   & 13.372    & 0.4750\\
\midrule
FLUX & Native RoPE          & 0.5665 & 31.217 & 16.631   & 0.7705\\
FLUX & No RoPE in injection & 0.0748 & \textbf{27.353} & 17.813   & 0.4540\\
FLUX & No RoPE globally     & \textbf{0.0002} & 27.897 & 18.533   & 0.4282\\
\bottomrule
\end{tabular}
\end{adjustbox}
\end{table}

\subsection{Effect of Affinity-Coherent Refinement}
\label{sec:acr_effect}
Removing the coordinate preference enlarges the set of style keys each query
can reach, but it also lets related content tokens settle on different keys.

\paragraph{Measuring retrieval coherence.}
We introduce \emph{Contrastive Style Evidence Consistency} (CSEC) to measure whether related content tokens retrieve more consistent style evidence than less-related tokens. For each content token $i$, cosine similarity between frozen content features identifies a related token $p$ and a less-related token $n$, both at the same spatial distance from $i$. At each sampled layer and step, we cluster style tokens into 32 groups using their frozen value features. Each content token's evidence is represented by a normalized histogram counting the groups of its eight highest-ranked retrieved style tokens. Let $e_i$, $e_p$, and $e_n$ denote these histograms for tokens $i$, $p$, and $n$, respectively:

\begin{equation}
\mathrm{CSEC}@8=100\,\mathbb{E}_{(i,p,n)} \left[\mathrm{JS}_2(e_i,e_n)-\mathrm{JS}_2(e_i,e_p)\right],
\label{eq:csec}
\end{equation}

where $\mathrm{JS}_2$ is base-2 Jensen--Shannon divergence and the expectation averages the selected triplets. Higher CSEC indicates stronger agreement for related pairs relative to less-related pairs. We also report \emph{Style Evidence Consistency} (SEC), $\mathrm{SEC}@8=100\bigl(1-\mathbb{E}_{(i,p,n)} \mathrm{JS}_2(e_i,e_p)\bigr)$, which measures agreement within related pairs.

\paragraph{Controlled comparison.}
We compare No ACR, spatial-neighbour coordination, random-neighbour coordination, and ACR. The three coordination conditions use the same operator, 16 neighbours, and self-inclusion rule. Only neighbour selection differs. All conditions share the same generation settings within each backbone.

Content-token pairs are selected using either cached content-inversion queries (Q) or frozen DINOv2-base features (DINO). Sampling and aggregation details are provided in the supplement.

\begin{table}[tb]
\centering
\caption{Retrieval coherence on FLUX and SDXL. Q and DINO identify the content features used to select evaluation pairs. Higher is better.}
\label{tab:acr_coherence}
\small
\renewcommand{\arraystretch}{0.9}
\setlength{\tabcolsep}{3pt}
\setlength{\aboverulesep}{1.5pt}
\setlength{\belowrulesep}{1.5pt}
\begin{adjustbox}{max width=\columnwidth}
\begin{tabular}{@{}lccc@{}}
\toprule
Condition & SEC@8 (Q)$\uparrow$ & CSEC@8 (Q)$\uparrow$ & CSEC@8 (DINO)$\uparrow$ \\
\midrule
\multicolumn{4}{l}{\emph{SDXL}}\\
No ACR & 58.12 & 37.06 & 11.22 \\
Spatial neighbours & 68.56 & 28.49 & 13.61 \\
Random neighbours & 60.02 & 0.67 & $-0.46$ \\
ACR & 67.70 & \textbf{45.78} & \textbf{14.46} \\
\midrule
\multicolumn{4}{l}{\emph{FLUX}}\\
No ACR & 56.64 & 20.54 & 9.27 \\
Spatial neighbours & 76.50 & 12.72 & 11.36 \\
Random neighbours & 78.68 & 0.39 & 0.08 \\
ACR & 71.25 & \textbf{29.22} & \textbf{14.77} \\
\bottomrule
\end{tabular}
\end{adjustbox}
\end{table}

\paragraph{Results.}
ACR achieves the highest mean CSEC under both feature definitions on both backbones (\cref{tab:acr_coherence}). CSEC(Q) rises from 20.54 to 29.22 on FLUX and from 37.06 to 45.78 on SDXL. Spatial and random coordination can increase SEC without comparable CSEC: on FLUX, random coordination reaches 78.68 SEC but only 0.39 CSEC. Thus, content affinity improves agreement specific to content relations beyond the agreement induced by spatial or random coordination.

\begin{figure}[tb]
    \centering
    \includegraphics[width=0.9\linewidth]{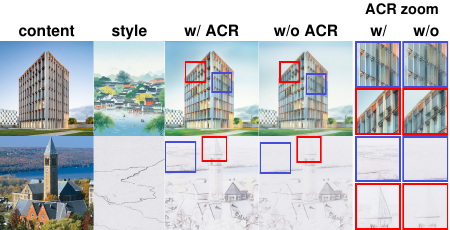}
    \caption{\textbf{Effect of affinity-coherent refinement (ACR).}
    Outputs generated with and without ACR under identical settings. The red
    and blue boxes mark the regions magnified in the two rightmost columns,
    where the differences are clearest. \textit{Please zoom in for better visulization.}}
    \label{fig:acr}
\end{figure}

\Cref{fig:acr} provides visual comparisons with and without ACR.

\begin{figure}[tb]
    \centering
    \includegraphics[width=0.82\linewidth]{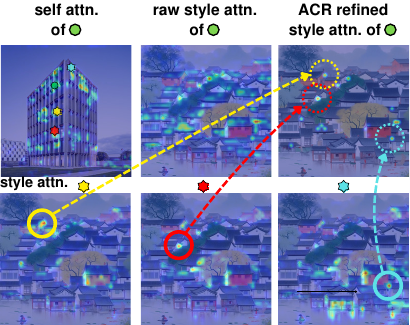}
    \caption{\textbf{How ACR refines a retrieval distribution.}
    Top: the content self-attention of the green query, its style attention
    before and after refinement.
    Bottom: the style attention of the three content tokens that the green
    query attends to most (yellow, red, cyan).
    ACR adds the dominant style peak of each neighbor (solid circles) to the
    candidate set of the green query, giving the consolidated peaks marked by
    the dotted circles above.}
    \label{fig:acr_attention}
\end{figure}

Without ACR the boxed regions degrade in two different ways: spurious mottled
texture appears on the flat wall surfaces in the first row, while in the second
row the pencil strokes of the reference are largely missing from the trees and
the tower.
Both are what incoherent retrieval predicts. Neighboring queries drawing on
incompatible style regions blotch a surface that should be uniform, and fail to
sustain a stroke pattern that should be continuous.
With ACR the same regions keep the texture of the reference.
\Cref{fig:acr_attention} shows the mechanism behind this on the first example.

\Cref{tab:components} separates the two contributions.
On SDXL, PDR and ACR each improve ArtFID over the injection baseline and are
better together than either alone.
On FLUX, PDR accounts for the large gain in content preservation, and ACR
further improves ArtFID and LPIPS at a small cost in FID.

\subsection{Comparison with Prior Methods}

\paragraph{Quantitative comparison.}
We compare against the following diffusion-based style transfer
methods: DiffStyle~\cite{Diffstyle}, DiffuseIT~\cite{diffuseit},
InstantStyle~\cite{instantstyle}, InstantStyle+~\cite{instantstyle_plus},
StyleAligned~\cite{StyleAlign}, StyleID~\cite{styleid},
StyleSSP~\cite{stylessp}, and StyleGallery~\cite{stylegallery}. We also compare against the optimization-based
InST~\cite{zhang2023inst} and, on the DiT side, USO~\cite{USO}.
\Cref{tab:sota} reports all three metrics under the 2,060-pair protocol.
With its SDXL backbone, StyleID++ obtains the lowest ArtFID and FID among the
U-Net-based methods, while StyleGallery achieves the lowest LPIPS.
On the DiT side, StyleID++ improves all three metrics over USO.
The two DiT rows are not a like-for-like ranking: USO is a trained
customization method, whereas StyleID++ is an inference-time intervention.
Read together with \cref{tab:leakage,tab:position_ablation}, these numbers show
that a single training-free intervention removes a failure mode shared by two
unrelated position mechanisms and improves output quality on both backbones.

\begin{table}[tb]
\centering
\caption{\textbf{Component ablation.} PDR and ACR provide complementary
improvements on SDXL. On FLUX, adding ACR to PDR improves ArtFID and LPIPS,
with a small FID trade-off.}
\label{tab:components}
\begin{adjustbox}{width=\columnwidth}
\begin{tabular}{llccc}
\toprule
Backbone & Configuration & ArtFID$\downarrow$ & FID$\downarrow$ & LPIPS$\downarrow$\\
\midrule
SDXL & Injection baseline & 22.489 & 14.190 & 0.4805\\
SDXL & + PDR & 21.198 & 13.372 & 0.4750\\
SDXL & + ACR & 21.054 & 13.265 & 0.4760\\
SDXL & + PDR + ACR & \textbf{20.894} & \textbf{13.197} & \textbf{0.4717}\\
\midrule
FLUX & Injection baseline & 31.217 & \textbf{16.631} & 0.7705\\
FLUX & + PDR & 27.353 & 17.813 & 0.4540\\
FLUX & + PDR + ACR & \textbf{26.280} & 18.047 & \textbf{0.3797}\\
\bottomrule
\end{tabular}
\end{adjustbox}
\end{table}

\begin{table}[tb]
\centering
\caption{Quantitative comparison with diffusion model baselines on all 2,060 content--style pairs.}
\label{tab:sota}
\begin{adjustbox}{max width=\columnwidth}
\begin{tabular}{lccc}
\toprule
Method & ArtFID$\downarrow$ & FID$\downarrow$ & LPIPS$\downarrow$\\
\midrule
\multicolumn{4}{l}{\emph{U-Net backbone}}\\
DiffStyle~\cite{Diffstyle} & 41.464 & 20.903 & 0.8931\\
InST~\cite{zhang2023inst} & 40.633 & 21.571 & 0.8002\\
DiffuseIT~\cite{diffuseit} & 40.721 & 23.065 & 0.6921\\
InstantStyle~\cite{instantstyle} & 37.524 & 21.817 & 0.6446\\
StyleAligned~\cite{StyleAlign} & 36.269 & 20.338 & 0.6997\\
InstantStyle+~\cite{instantstyle_plus} & 25.886 & 16.097 & 0.5140\\
StyleID~\cite{styleid} & 23.172 & 14.400 & 0.5049\\
StyleSSP~\cite{stylessp} & 22.060 & 13.854 & 0.4851\\
StyleGallery~\cite{stylegallery} & 21.495 & 14.110 & \textbf{0.4226}\\
StyleID++ (Ours) & \textbf{20.894} & \textbf{13.197} & 0.4717\\
\midrule
\multicolumn{4}{l}{\emph{DiT backbone}}\\
USO~\cite{USO} & 29.453 & 19.000 & 0.4727\\
StyleID++ (Ours) & \textbf{26.280} & \textbf{18.047} & \textbf{0.3797}\\
\bottomrule
\end{tabular}
\end{adjustbox}
\end{table}

\begin{table}[t]
\centering
\caption{User-study results.}
\label{tab:user_study}
\begin{adjustbox}{width=\columnwidth}
\begin{tabular}{lccc}
\toprule
Method & Content$\uparrow$ & Style$\uparrow$ & Overall$\uparrow$\\
\midrule
StyleID & $3.34{\pm}0.36$ & $3.85{\pm}0.18$ & $3.65{\pm}0.33$\\
StyleSSP & $3.74{\pm}0.27$ & $3.58{\pm}0.65$ & $3.70{\pm}0.20$\\
StyleID++ & $\mathbf{4.08{\pm}0.29}$ & $\mathbf{4.15{\pm}0.22}$ &
$\mathbf{4.08{\pm}0.20}$\\
\bottomrule
\end{tabular}
\end{adjustbox}
\end{table}

\begin{figure*}[tb]
    \centering
    \includegraphics[width=0.96\textwidth]{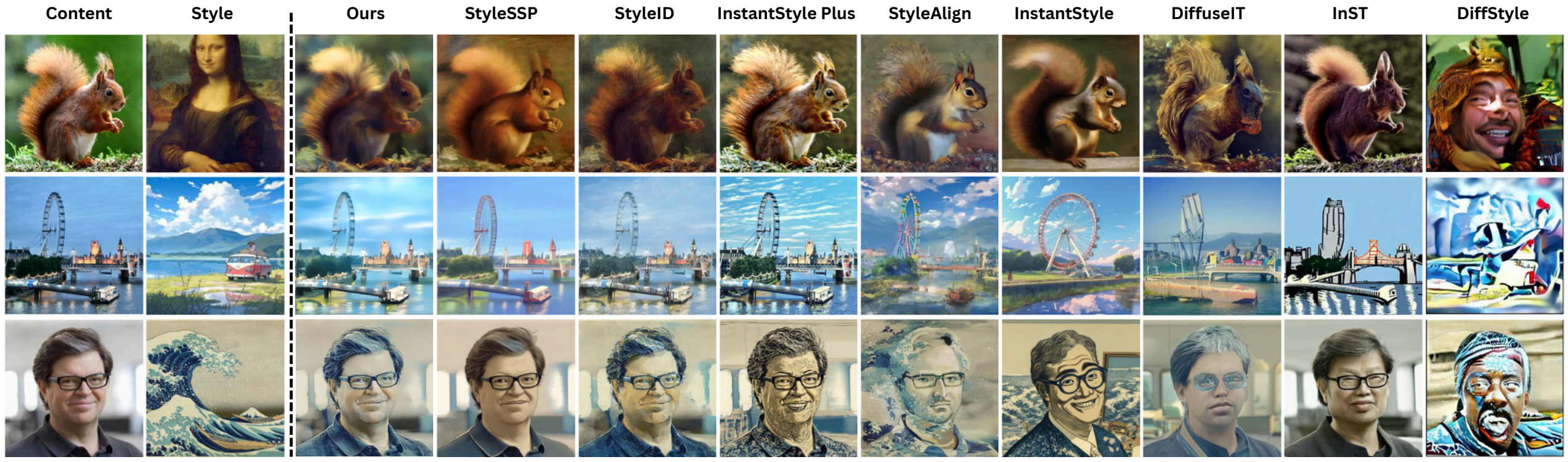}
    \caption{Broader comparison with diffusion-based style transfer
methods. \textit{Please zoom in for better visualization.} \textit{Additional results are provided in the supplementary material.}}
    \label{fig:comparison}
\end{figure*}

\begin{figure}[tb]
    \centering
    \includegraphics[width=\linewidth]{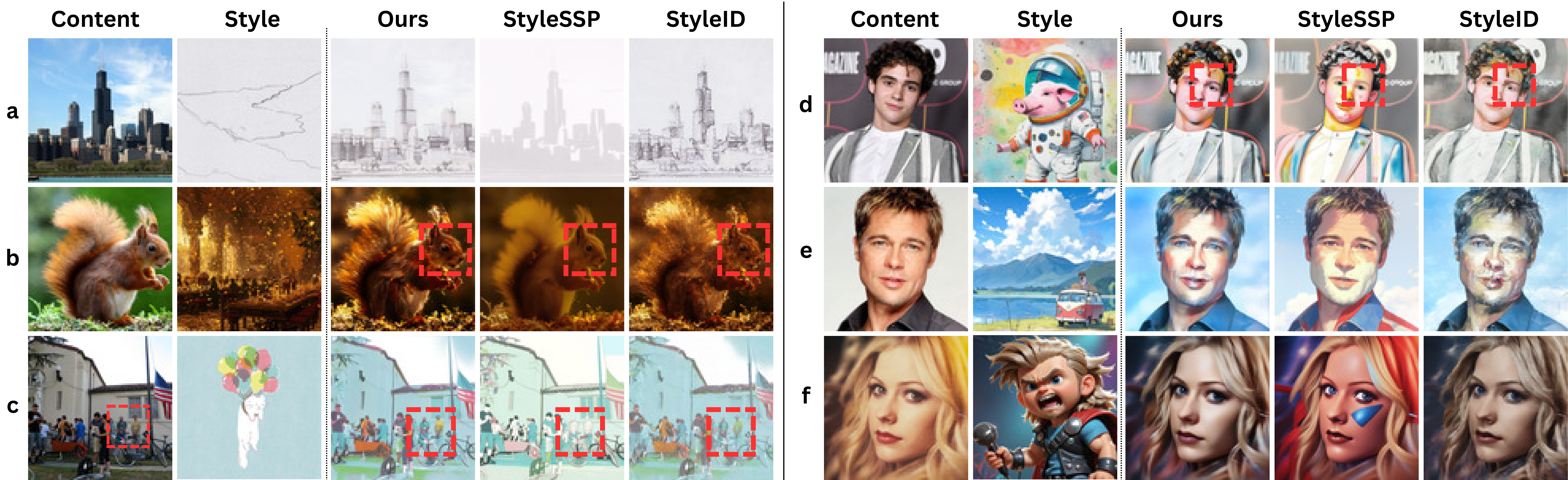}
    \caption{\textbf{Comparison with StyleSSP and StyleID on difficult pairs.}
Panels (a)--(c) pair photographs with references whose character is carried
by fine texture. Panels (d)--(f) pair portraits with strongly stylized cartoon
references. Red dashed boxes mark corresponding regions for close
comparison. \textit{Please zoom in for better visualization.}}
    \label{fig:comparison_hard}
\end{figure}

\paragraph{Qualitative comparison.}
\Cref{fig:comparison} compares diffusion baselines, while
\cref{fig:comparison_hard} examines harder pairs with StyleSSP and StyleID.
StyleAligned, InstantStyle, InST, and DiffuseIT achieve strong stylization
by altering content identity, including the squirrel's pose and the
portrait's face. InstantStyle+ and StyleSSP preserve structure, but
StyleSSP's smooth, low-detail textures can leave stylization incomplete.
StyleID preserves layout but coarsens fine detail. These limitations become
clearer in \cref{fig:comparison_hard}: StyleSSP almost ignores reference
contours and textures in (a) and (b) and leaks style-image content in
(c)--(f), while StyleID retains layout across (a)--(e) but coarsens detail.
StyleID++ preserves both layout and fine detail while transferring reference
texture. \textit{More results appear in the supplementary material.}

\paragraph{User study.}
We further conducted a \textbf{user study} with 20 participants over 30 content–style pairs. Each result was rated on a 1(poor)–5(excellent) Likert scale along three dimensions: content preservation, style fidelity, and overall preference.
As shown in \Cref{tab:user_study}, StyleID++ achieves the highest average
scores (Content: \textbf{4.08}, style: \textbf{4.15}, overall: \textbf{4.08}),
outperforming StyleID (3.34 / 3.85 / 3.65) and StyleSSP (3.74 / 3.58 / 3.70).
Paired Wilcoxon signed-rank tests show that StyleID++ significantly outperforms
both baselines on all three dimensions ($p < 0.01$), consistent with our
quantitative improvements in FID and LPIPS.

\section{Conclusion}
Viewing attention injection as cross-image retrieval exposes how useful
within-image position priors become coordinate shortcuts between unrelated
layouts. This occurs in both a convolutional U-Net and a DiT despite their
different position mechanisms, implicating the reuse of within-image priors
rather than a single operator. Removing the shortcut is necessary but
insufficient: unconstrained retrieval still lets related tokens select
incompatible style evidence. StyleID++ therefore combines two training-free
components: PDR suppresses coordinate preference, and ACR uses content
self-attention to coordinate related queries. Coordinate--centroid
correlation and a permutation test measure the shortcut directly, and both
components improve ArtFID on SDXL and FLUX. The same mismatch motivates
examining position decoupling wherever frozen within-image attention is
reused between independently composed images in other training-free
transfer settings.

{
    \small
    \bibliographystyle{ieeenat_fullname}
    \bibliography{main}
}

\appendix
\setcounter{figure}{0}
\setcounter{table}{0}
\setcounter{equation}{0}
\renewcommand{\thefigure}{A\arabic{figure}}
\renewcommand{\thetable}{A\arabic{table}}
\renewcommand{\theequation}{A\arabic{equation}}
\renewcommand{\theHfigure}{A\arabic{figure}}
\renewcommand{\theHtable}{A\arabic{table}}
\renewcommand{\theHequation}{A\arabic{equation}}
\setcounter{secnumdepth}{3}
\setlength{\tabcolsep}{5pt}
\renewcommand{\topfraction}{0.9}
\renewcommand{\bottomfraction}{0.8}
\renewcommand{\textfraction}{0.07}
\renewcommand{\floatpagefraction}{0.75}
\renewcommand{\dbltopfraction}{0.9}
\renewcommand{\dblfloatpagefraction}{0.75}
\setcounter{topnumber}{3}
\setcounter{dbltopnumber}{3}
\setcounter{totalnumber}{5}

\maketitlesupplementary

\section*{Overview}
This supplementary material provides additional visualizations, hyperparameter
configurations, and runtime analyses to complement the results presented in
the main paper. The sections are organized as follows:
\begin{itemize}
    \item \textbf{\Cref{sec:more_visualizations}}: Additional qualitative results
    \item \textbf{\Cref{sec:pdr_attention}}: Image-level swap comparisons
        and attention-level analysis of positional leakage and PDR on both
        backbones.
    \item \textbf{\Cref{sec:hyperparams}}: Hyperparameter configurations used
        in our experiments and their effect on the evaluation metrics.
    \item \textbf{\Cref{sec:runtime}}: Evaluation of runtime and memory usage.
    \item \textbf{\Cref{sec:csec_protocol}}: Evaluation settings for the
        CSEC retrieval-coherence diagnostic.
\end{itemize}

\section{Additional Visualizations}
\label{sec:more_visualizations}

In this section, we provide more qualitative results on diverse content--style
pairs to complement those in the main paper.

\begin{figure*}[tp]
    \centering
    \includegraphics[width=\linewidth]{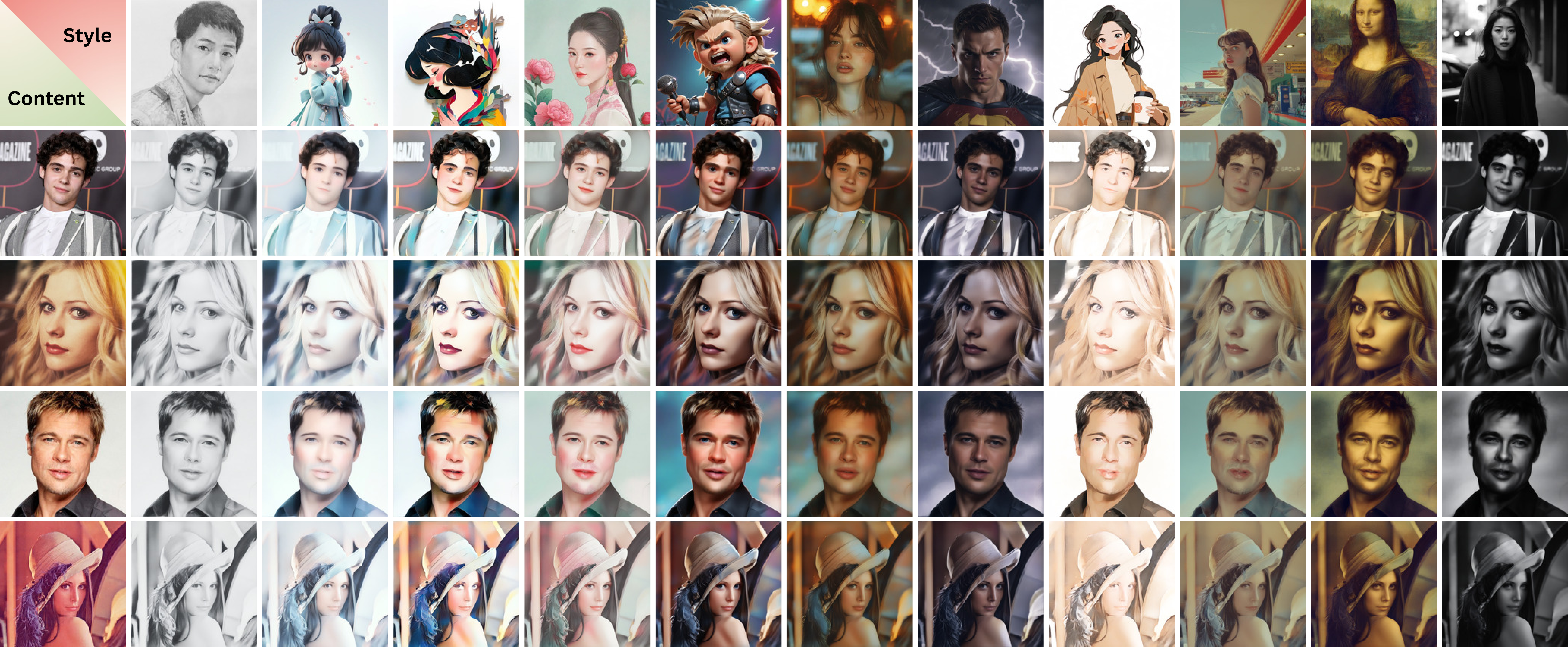}
    \caption{\textbf{Portrait-to-portrait stylization results produced by our method.}
    Our method effectively transfers fine facial textures, color tones, and
    shading patterns, while maintaining the identity and structural
    consistency of the content portrait. \textit{Please zoom in to examine
    transferred textures and fine details.}}
    \label{fig:ours_portrait}
\end{figure*}

\begin{figure*}[tp]
    \centering
    \includegraphics[width=\linewidth]{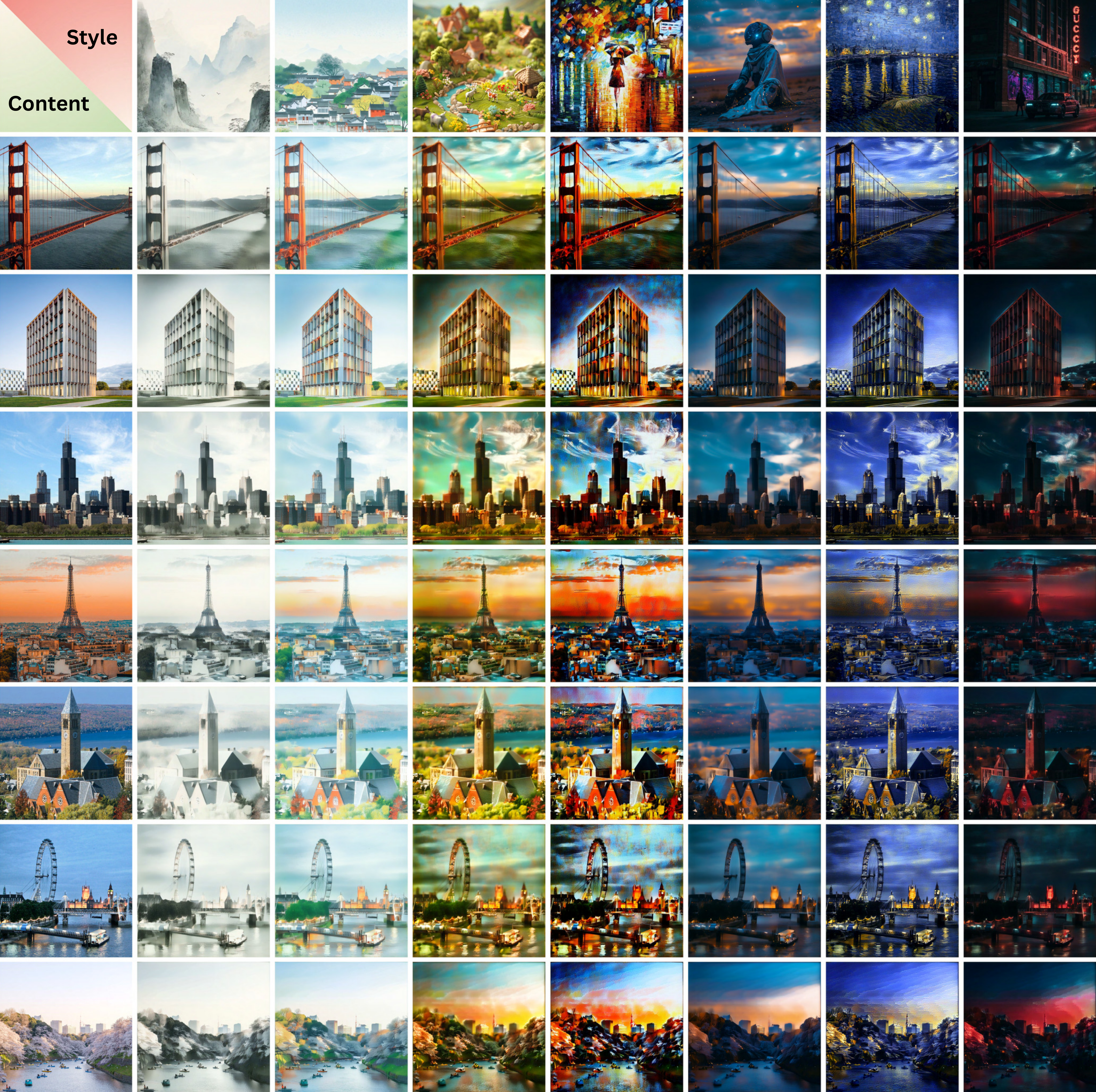}
    \caption{\textbf{Landscape-to-landscape stylization results.}
    Our method successfully transfers global color distributions, lighting
    conditions, and spatial textures while preserving scene layout, object
    boundaries, and overall geometry. \textit{Please zoom in to examine fine
    details.}}
    \label{fig:ours_landscape}
\end{figure*}

\begin{figure*}[tp]
    \centering
    \includegraphics[width=\linewidth]{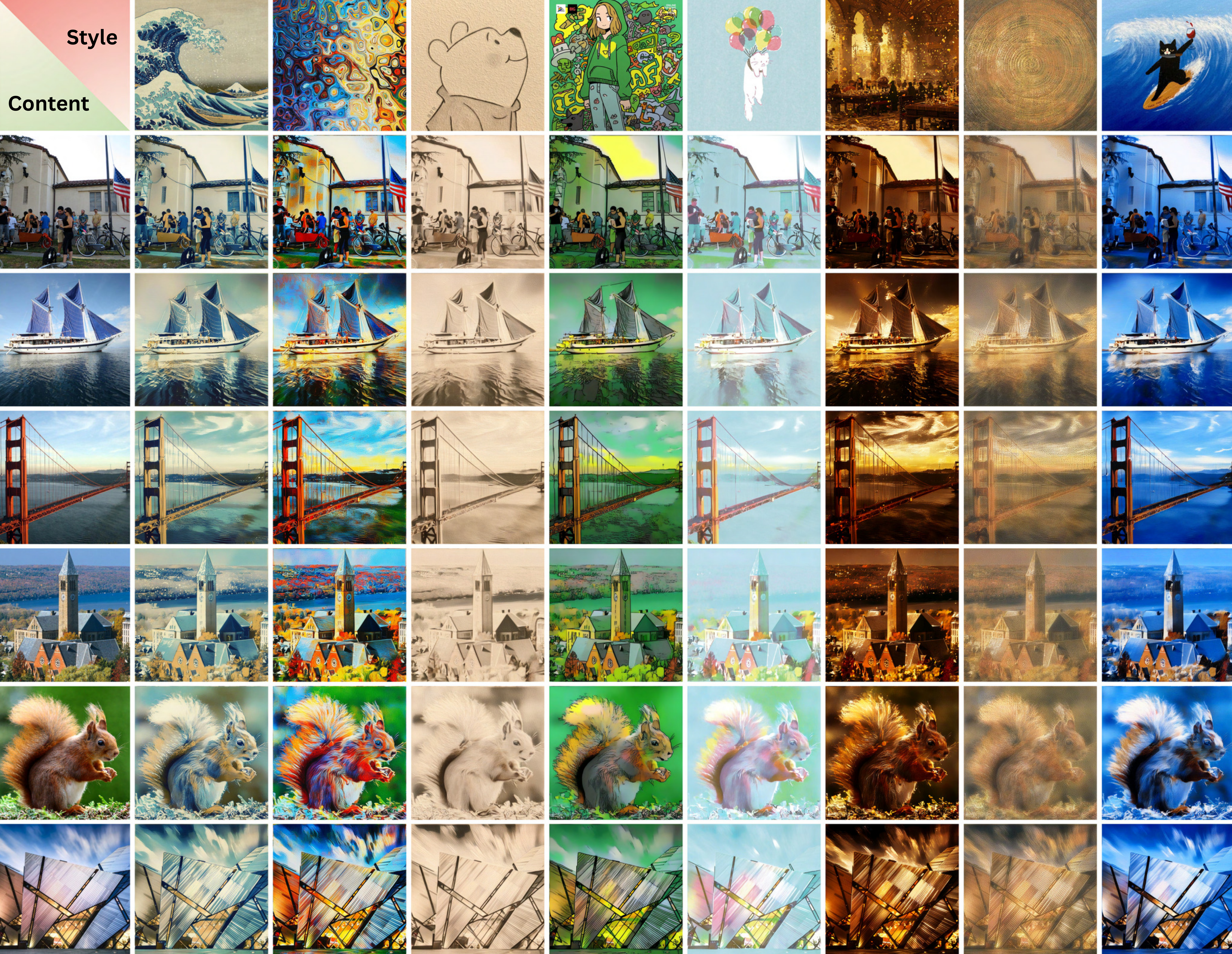}
    \caption{\textbf{Cross-domain stylization with challenging style elements.}
    This figure includes content--style pairs from different semantic domains,
    where the style images exhibit strong textures, abstract patterns, or
    other challenging high-frequency elements. Our method remains robust in
    these cases, producing coherent stylization results without introducing
    structural artifacts. \textit{Please zoom in to examine fine details.}}
    \label{fig:ours_crossdomain}
\end{figure*}

\begin{figure*}[tp]
    \centering
    \includegraphics[width=\linewidth]{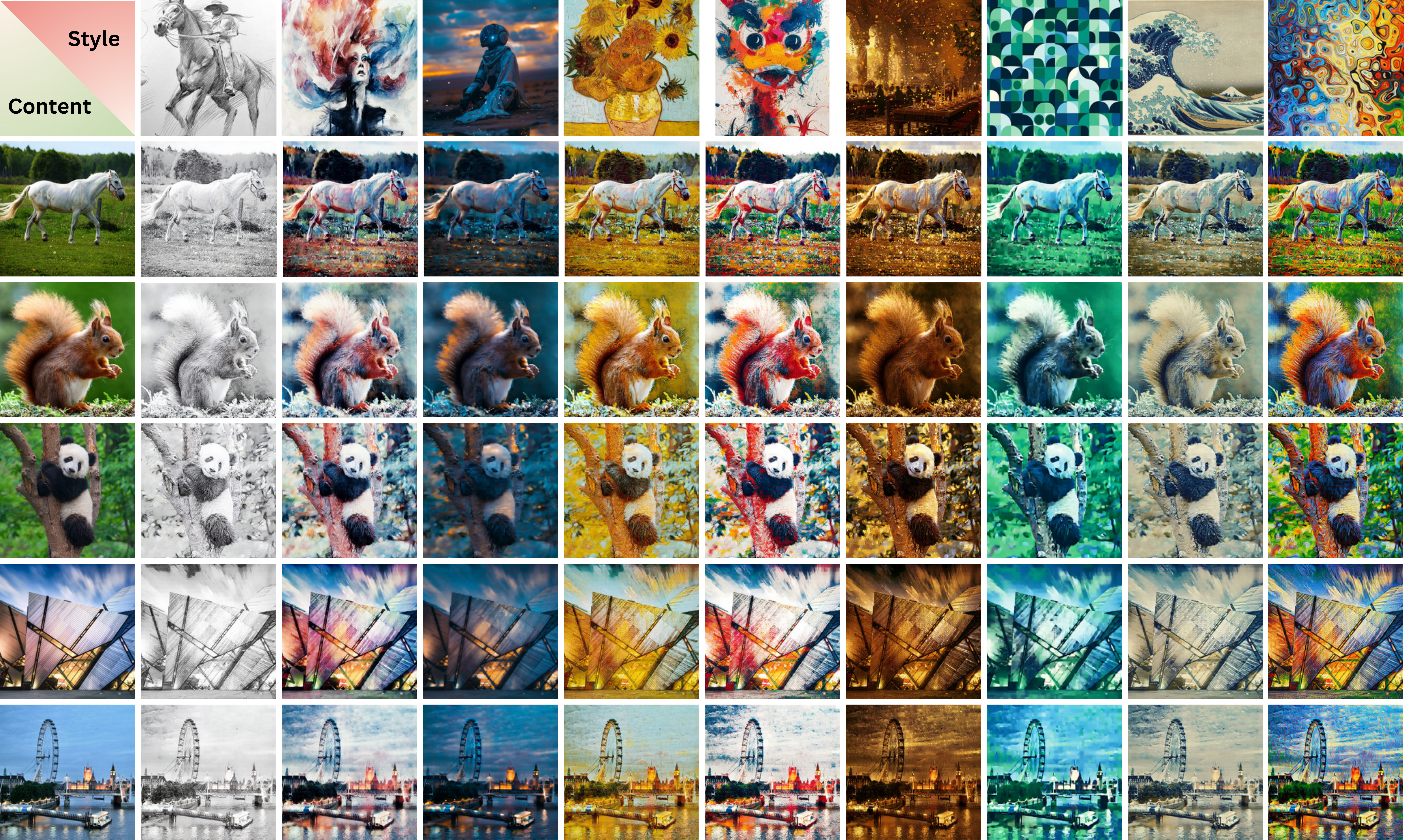}
    \caption{\textbf{More results generated by StyleID++ on FLUX backbone}
    This figure shows cross-domain style transfer results generated by
    StyleID++ on the FLUX backbone. It demonstrates that our method remains
    robust on the DiT architecture. \textit{Please zoom in to examine
    details.}}
    \label{fig:ours_flux}
\end{figure*}

\begin{figure*}[tp]
    \centering
    \includegraphics[width=\linewidth]{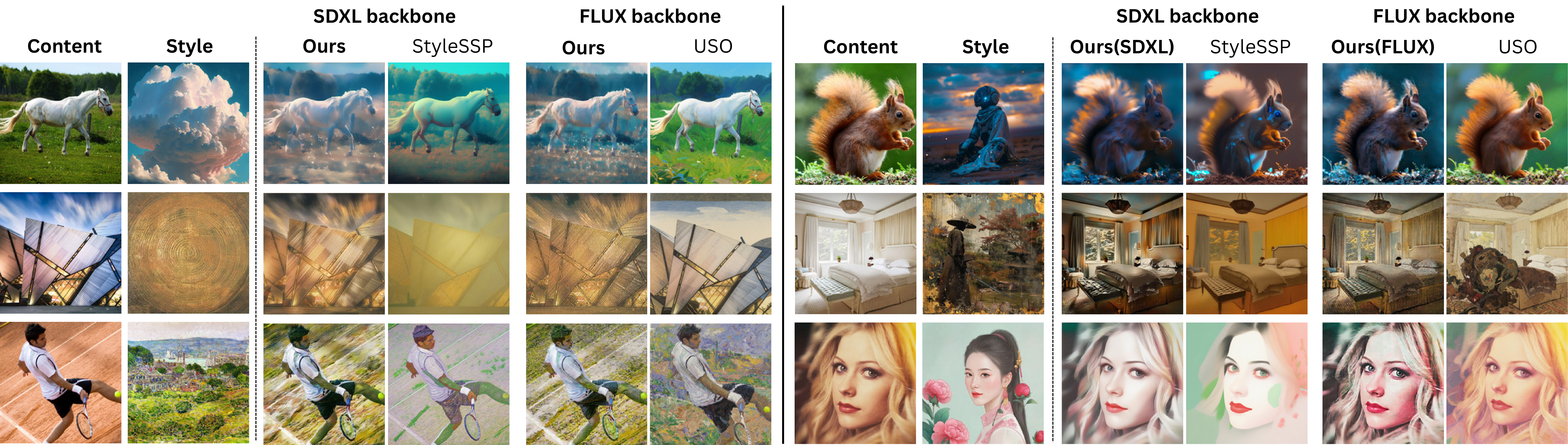}
   \caption{\textbf{Qualitative comparison of StyleID++, StyleSSP, and USO}.
       Our method demonstrates robust content structure preservation and
       effectively prevents content leakage from the style image.}

    \label{fig:compare3}
\end{figure*}

\Cref{fig:ours_portrait,fig:ours_landscape,fig:ours_crossdomain} present
additional stylization results of our method across three categories: (1)
portrait-to-portrait stylization, (2) landscape-to-landscape stylization, and
(3) cross-domain stylization. These examples cover visually diverse
content--style combinations, including challenging style patterns with strong
textures, irregular brush strokes, or geometric distortions. Across all
categories, our method consistently preserves semantic structures while
capturing both global style appearance and fine-grained high-frequency
details. \Cref{fig:ours_flux} shows results on the FLUX backbone, and
\Cref{fig:compare3} compares StyleID++ with StyleSSP and USO~\cite{USO}, where
our method preserves the content structure and avoids leaking content from the
style image.

\section{Positional Leakage under Style Rearrangements}
\label{sec:pdr_attention}

\begin{figure*}[p]
    \centering
    \includegraphics[width=\textwidth]{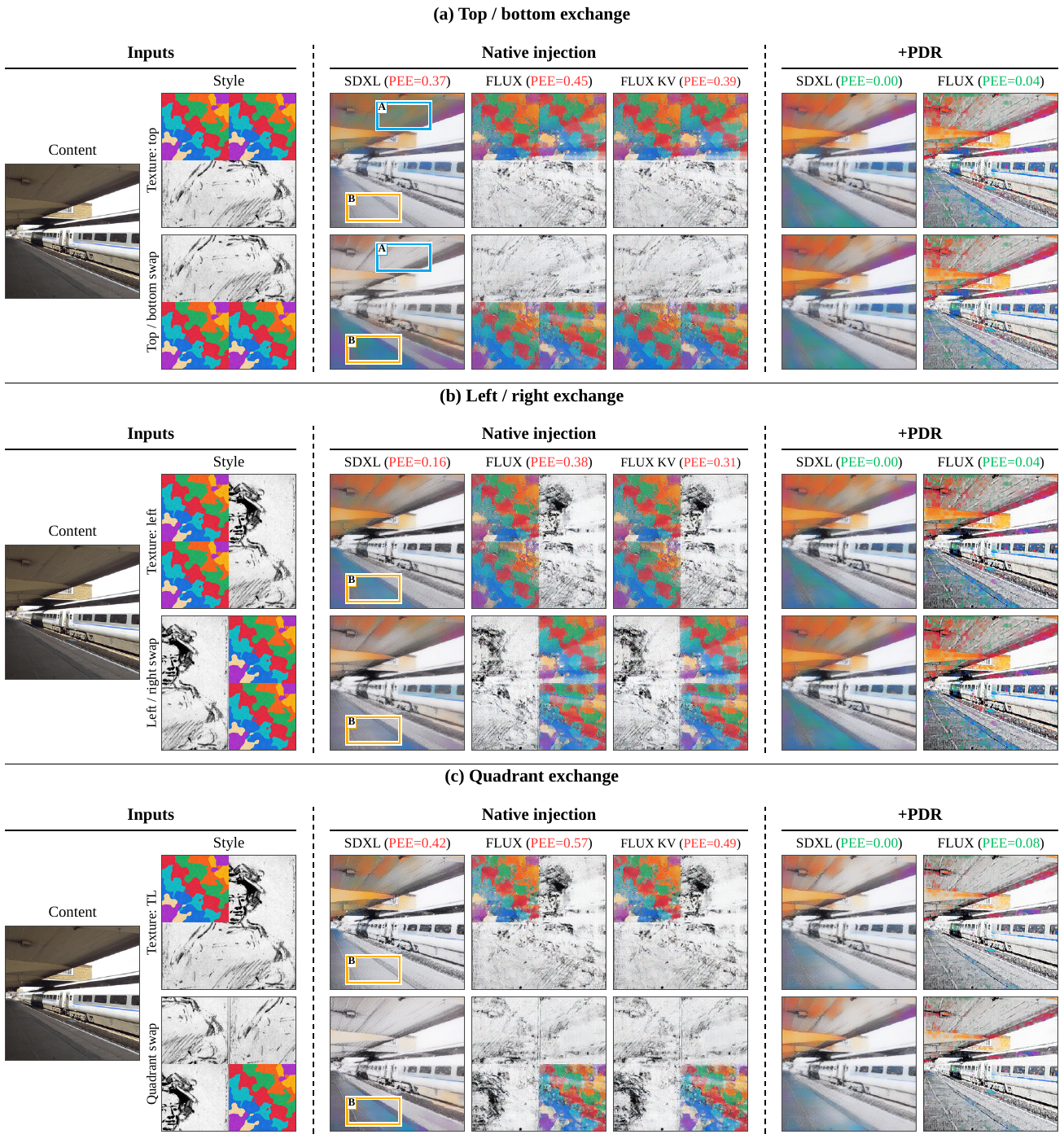}
    \caption{\textbf{Positional leakage under three style rearrangements.}
    Panels (a)--(c) show top/bottom, left/right, and quadrant exchanges,
    respectively. Each panel compares native SDXL, FLUX, and FLUX KV
    injection with PDR on SDXL and FLUX. FLUX KV permutes feature token
    positions from the original style image instead of swapping pixel regions.
    In the native SDXL column,
    matching A/B labels mark the same content regions before and after
    each swap. The displayed PEE values describe the individual examples.
    PDR reduces sensitivity to the rearrangement across all three cases.}
    \label{fig:three_swaps}
\end{figure*}

\paragraph{Image-level comparisons.}
\Cref{fig:three_swaps} extends the example in Figure~1 of the main paper
to three rearrangements of the style reference: top/bottom, left/right,
and quadrant exchanges. Each pair of rows keeps the content image fixed.
SDXL and FLUX exchange regions of the style image. FLUX KV instead applies
the corresponding permutation to feature token positions from the original
style image. In the native SDXL outputs,
the marked roof (A) and platform (B) regions illustrate how the color
response changes with texture position. FLUX and FLUX KV more strongly
reproduce the reference layout. Across these examples, PDR yields more
consistent outputs while preserving the content structure.

\begin{figure*}[tp]
    \centering
    \includegraphics[width=\linewidth]{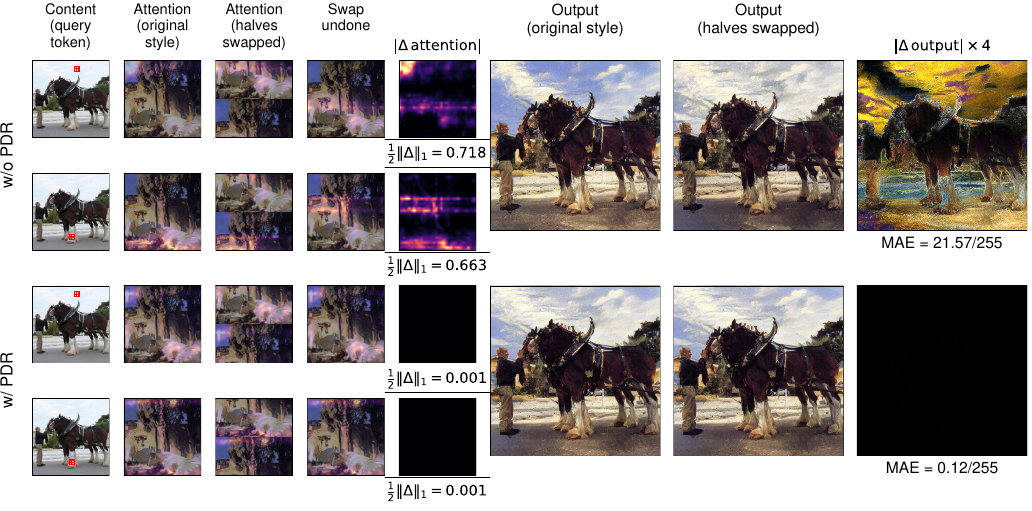}
    \caption{\textbf{Position-anchored attention on SDXL, and its removal by PDR.}
    For each content query (red box) we show its attention over the style keys
    before and after the two halves of the style image are exchanged, the same
    attention aligned back by undoing the known permutation, and the residual
    between the two. The output columns do not depend on the query and are
    therefore shared by the two rows of each block. Without PDR both queries
    retrieve by coordinate rather than by appearance, but in different ways:
    the sky query (first row) follows the position, while the hoof query
    (second row) follows the padding artefacts at the image boundary. With PDR
    both maps translate exactly with the swapped halves, the residual
    vanishes, and the output is unchanged pixel by pixel
    ($\mathrm{MAE}=0.12/255$ against $21.57/255$).}
    \label{fig:pdr_attn_sdxl}
\end{figure*}

\begin{figure*}[tp]
    \centering
    \includegraphics[width=\linewidth]{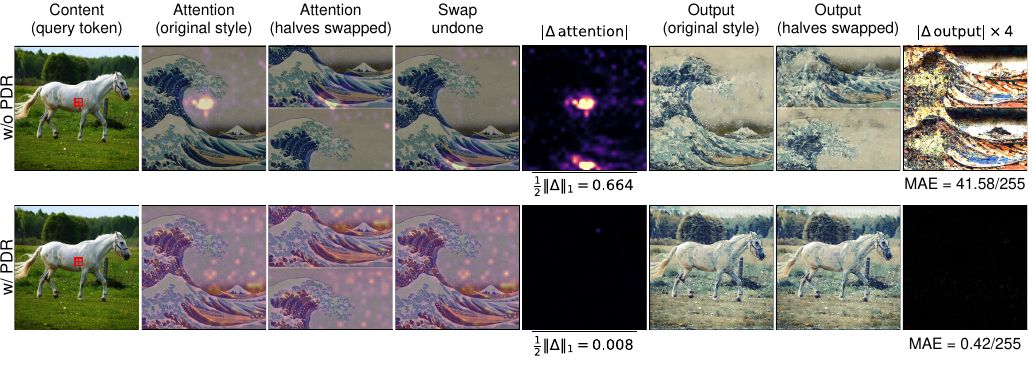}
    \caption{\textbf{Position-anchored attention on FLUX, and its removal by PDR.}
    Columns are as in \Cref{fig:pdr_attn_sdxl}. Without PDR the peak stays at
    the same absolute position after the swap, so undoing the swap displaces
    it by half the image height and leaves a double-blob residual, and the
    output changes substantially ($\mathrm{MAE}=41.58/255$). With PDR the map
    translates exactly with the swapped halves, the residual falls to the bf16
    noise floor, and the output is unchanged pixel by pixel
    ($\mathrm{MAE}=0.42/255$).}
    \label{fig:pdr_attn_flux}
\end{figure*}

\paragraph{Attention-level analysis.}
\Cref{fig:pdr_attn_sdxl,fig:pdr_attn_flux} apply the permutation test of the
main paper to a single content query on each backbone.

Without PDR the two backbones anchor on coordinates in different ways. On SDXL
the sky query keeps reading the top rows of the style image: before the swap
those rows hold the sky it is looking for, afterwards they hold a white cow,
so the attended position is unchanged while its content has been replaced. The
query at the horse's hoof instead locks onto the boundary artefacts of zero
padding, and its attention hugs horizontal bands at fixed absolute rows.
Undoing the swap displaces those bands to the middle seam and leaves a striped
residual. On FLUX a single peak stays at the same absolute position, so
undoing the swap displaces it by half the image height.

In these attention-level examples, PDR makes the attention translate exactly
with the swapped halves on both backbones, and the residual falls to the
numerical noise floor. The
equivariance carries through to the image: exchanging the halves of the style
reference leaves the generated result unchanged pixel by pixel, and the small
non-zero MAE reported in the figures is floating-point accumulation rather
than residual coordinate dependence.

\begin{figure}[!htbp]
    \centering
    \includegraphics[width=0.9\linewidth]{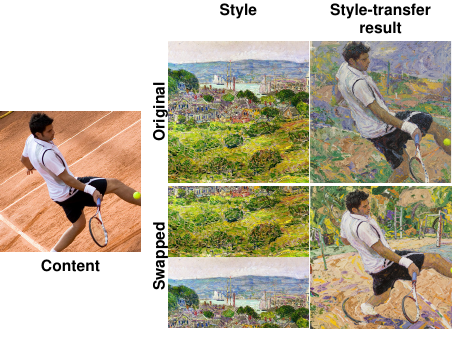}
    \caption{\textbf{Positional leakage in a trained DiT method.}
    Results of USO for the same content image with the original and the
    vertically swapped style reference. When the green field moves to the
    upper half of the reference, the green region of the output moves with it,
    so the output layout tracks the reference rather than the structure of the
    content.}
    \label{fig:dit_bias}
\end{figure}

Positional leakage is not confined to training-free injection.
\Cref{fig:dit_bias} applies the same swap to USO, a trained DiT model for
reference-driven generation, and its output layout follows the layout of the
style reference.
Methods that are not training-free are outside the scope of this work, so we
report this only as an observation.

\section{Hyperparameter Configurations}
\label{sec:hyperparams}

Following StyleID~\cite{styleid}, injection is applied at every denoising step
rather than being restricted to a sub-interval of the sampling trajectory. On
SDXL it is applied to the upsampling blocks 3, 4 and 5 of the U-Net. On FLUX
it is applied to the double-stream blocks 12--18 and to the single-stream
blocks 0--37.

\Cref{tab:hyperparams_k,tab:hyperparams_lambda} present the effect of ACR
hyperparameters on evaluation metrics. $\kappa_q$ controls the number of
top-attended neighbors selected from the content self-attention map.
$\kappa_s$ specifies the number of peak keys retained in the style attention
for the query itself. Each neighbor contributes only its single dominant key.
$\lambda_{\mathrm{self}}$ and $\lambda_{\mathrm{aff}}$ control the strength of
influence that the self and neighbor peaks exert on the refined logits. On
SDXL our default configuration is $\kappa_q=8$, $\kappa_s=16$,
$\lambda_{\mathrm{self}}=0.3$, and $\lambda_{\mathrm{aff}}=0.1$
(\Cref{tab:hyperparams_lambda,tab:hyperparams_k}).
\Cref{tab:hyperparams_lambda_flux} reports the same sweep on the FLUX
backbone, where the default configuration is $\kappa_q=256$, $\kappa_s=256$,
$\lambda_{\mathrm{self}}=0.2$, and $\lambda_{\mathrm{aff}}=0.2$. The two
backbones are tuned independently. Each sweep varies one factor around its own
default.

\begin{table}[!htbp]
\centering
\caption{Effect of $\lambda_{\mathrm{self}}$ and $\lambda_{\mathrm{aff}}$ on SDXL.
The default configuration used in our experiment is highlighted in bold.}
\label{tab:hyperparams_lambda}
\begin{tabular}{c|c|ccc}
\toprule
\textbf{$\lambda_{\mathrm{self}}$} & \textbf{$\lambda_{\mathrm{aff}}$}
& \textbf{ArtFID} $\downarrow$
& \textbf{FID} $\downarrow$
& \textbf{LPIPS} $\downarrow$ \\
\midrule

0.10 & 0.10 & 20.947 & 13.221 & 0.473 \\
0.10 & 0.15 & 21.064 & 13.310 & 0.472 \\
0.10 & 0.20 & 21.126 & 13.352 & 0.472 \\

\midrule

0.20 & 0.10 & 20.934 & 13.221 & 0.472 \\
0.20 & 0.15 & 21.014 & 13.286 & 0.471 \\
0.20 & 0.20 & 21.102 & 13.345 & 0.471 \\

\midrule

\textbf{0.30} & \textbf{0.10} & \textbf{20.894} & \textbf{13.197} & \textbf{0.472} \\
0.30 & 0.15 & 21.006 & 13.280 & 0.471 \\
0.30 & 0.20 & 21.081 & 13.341 & 0.470 \\

\bottomrule
\end{tabular}
\end{table}

\begin{table}[!htbp]
\centering
\caption{Effect of $\kappa_q$ and $\kappa_s$ on SDXL.
The default configuration used in our experiment is highlighted in bold.}
\label{tab:hyperparams_k}
\begin{tabular}{l|c|ccc}
\toprule
\textbf{Parameter} & \textbf{Value} & \textbf{ArtFID} $\downarrow$ & \textbf{FID} $\downarrow$ & \textbf{LPIPS} $\downarrow$ \\
\midrule

$\kappa_q$ & \textbf{8}  & \textbf{20.894} & \textbf{13.197} & \textbf{0.472} \\
      & 16 & 20.952 & 13.251 & 0.470 \\
      & 24 & 21.014 & 13.315 & 0.468 \\

\midrule

$\kappa_s$  & 8 & 20.945 & 13.258 & 0.469 \\
            & \textbf{16} & \textbf{20.894} & \textbf{13.197} & \textbf{0.472} \\
            & 24 & 20.882 & 13.166 & 0.474 \\

\bottomrule
\end{tabular}
\end{table}

\begin{table}[!htbp]
\centering
\caption{Effect of $\kappa_q$, $\kappa_s$, $\lambda_{\mathrm{self}}$ and
$\lambda_{\mathrm{aff}}$ on FLUX.
The default configuration used in our experiment is highlighted in bold.}
\label{tab:hyperparams_lambda_flux}
\begin{tabular}{c|c|c|ccc}
\toprule
$\kappa_q / \kappa_s$ & \textbf{$\lambda_{\mathrm{self}}$} & \textbf{$\lambda_{\mathrm{aff}}$}
& \textbf{ArtFID} $\downarrow$
& \textbf{FID} $\downarrow$
& \textbf{LPIPS} $\downarrow$ \\
\midrule

256 / 256   & 0.10 & 0.10 & 27.908 & 19.581 & 0.356 \\
            & 0.10 & 0.20 & 26.401 & 18.117 & 0.381 \\

            & 0.20 & 0.10 & 27.947 & 19.625 & 0.355 \\
            & \textbf{0.20} & \textbf{0.20} & \textbf{26.280} & \textbf{18.047} & \textbf{0.380} \\

            & 0.30 & 0.10 & 28.151 & 19.760 & 0.356 \\
            & 0.30 & 0.20 & 26.404 & 18.119 & 0.381 \\
 \midrule
 128 / 128  & 0.10 & 0.10 & 28.662 & 20.075 & 0.360 \\
            & 0.10 & 0.20 & 26.718 & 18.602 & 0.363 \\
            & 0.20 & 0.10 & 28.674 & 20.084 & 0.360 \\
            & 0.20 & 0.20 & 26.571 & 18.495 & 0.363 \\
            & 0.30 & 0.10 & 28.686 & 20.093 & 0.360 \\
            & 0.30 & 0.20 & 26.747 & 18.624 & 0.363 \\

\bottomrule
\end{tabular}
\end{table}

\section{Runtime Analysis}
\label{sec:runtime}

\Cref{tab:runtime} reports inference time, peak memory, and parameter count,
with the overhead column measured against StyleID. Every method is run at its
official default resolution and number of denoising steps, both listed in the
table, so the overhead column is not a like-for-like measure of algorithmic
cost. On the U-Net backbone StyleID++ takes $5.5$\,s per image against
$2.8$\,s for StyleID, roughly a factor of two, while producing $1024^2$ images
in $20$ steps against StyleID's $512^2$ in $50$. Compared with StyleSSP, the
strongest training-free U-Net baseline, StyleID++ is favourable in every
column: $5.5$\,s against $23.5$\,s ($4.3\times$ faster), $22.87$\,GB against $41.80$\,GB, and
$3.47$\,B parameters against $18.99$\,B.

\begin{table*}[!tp]
\centering
\caption{Default resolution and denoising steps, runtime, overhead (OH) with
respect to StyleID, peak memory, and parameter count.}
\label{tab:runtime}
\begin{tabular}{lcccccc}
\toprule
Method & Res. & Steps & Time (s) & OH & Mem (GB) & Params (B)  \\
\midrule
\multicolumn{7}{l}{\emph{U-Net backbone}}\\
StyleID(CVPR'23)\cite{styleid}    & $512^2$  & 50 & 2.8  & --        & 8.32  & 1.07  \\
StyleSSP(CVPR'24)\cite{stylessp}  & $1024^2$ & 50 & 23.5 & +739.3\%  & 41.80 & 18.99 \\
\textbf{StyleID++(SDXL)}          & $1024^2$ & 20 & 5.5  & +96.4\%   & 22.87 & 3.47  \\
\midrule
\multicolumn{7}{l}{\emph{DiT backbone}}\\
USO(CVPR'25)\cite{USO}            & $1024^2$ & 25 & 14.2 & +407.1\%  & 44.20 & 17.55 \\
\textbf{StyleID++(FLUX)}          & $1024^2$ & 14 & 13.5 & +382.1\%  & 48.01 & 12.11 \\
\bottomrule
\end{tabular}
\end{table*}

On the DiT backbone StyleID++ is slightly faster than USO ($13.5$\,s against
$14.2$\,s) and uses fewer parameters ($12.11$\,B against $17.55$\,B), but it
needs more peak memory ($48.01$\,GB against $44.20$\,GB). On both backbones
the overhead over StyleID comes from two sources: the size of the backbone we
build on, and the inversion, whose intermediate results have to be retained for
injection during sampling.
We modify only the attention logits and reuse the pretrained VAE and backbone,
so the method introduces no trainable parameters and no architectural change.
PDR replaces the padding rule and removes a positional encoding, and ACR
selects and reweights entries of the attention logits that are already
computed, so neither changes the time complexity of the backbone.

\section{CSEC Evaluation Settings}
\label{sec:csec_protocol}

We record each condition's actual retrieval at step indices 2, 10, and 18 (seed 0). For FLUX, we sample heads 0, 6, 12, and 18 in \texttt{double\_12}, \texttt{double\_18}, \texttt{single\_19}, and \texttt{single\_37}. For SDXL, we sample heads 0, 2, 5, and 7 in both transformer modules of injection layers 3--5. All indices start at zero. For Q-based relations, we concatenate the cached content-inversion queries from heads other than those evaluated and normalize each token's descriptor. Style groups use cosine $k$-means on the corresponding normalized style-value descriptors. DINO-based relations use frozen DINOv2-base~\cite{oquab2023dinov2} patch features aligned to the content-token grid by bilinear interpolation. Style groups and evaluation pairs are shared across conditions.

For each token $i$, related candidates lie within four grid units and have cosine similarity above its median similarity to all content tokens. We exclude the union of its top-16 neighbours under query cosine similarity, head-averaged content self-attention, and DINO cosine similarity, then retain up to four highest-similarity candidates. Each is paired with the most similar token at the same spatial distance whose similarity is at or below the median. Unmatched pairs are omitted. Scores are computed per head, then averaged equally over heads, valid layer--step combinations within each input pair, and input pairs. Empty evaluation positions are omitted for all conditions.

\FloatBarrier

\end{document}